\documentclass{article}
\usepackage{spconf,amsmath,graphicx}

\usepackage{enumitem}
\setlist{nosep, leftmargin=14pt}

\usepackage{mwe} 
\usepackage{threeparttable}

\usepackage{xcolor}
\definecolor{citecolor}{HTML}{0071BC}
\definecolor{linkcolor}{HTML}{ED1C24}
\usepackage[colorlinks,
            anchorcolor=red,
            citecolor=citecolor, 
            linkcolor=linkcolor,
            ]{hyperref}
\usepackage{multicol}

\usepackage{algorithm}
\usepackage{algorithmic}

\def\OurDatasetAbdomenAtlas{{AtlasBench}}
\def\OurDatasetJHH{{JHHBench}}

\newcommand{\pipeline}{{\fontfamily{ppl}\selectfont
Label Critic}}

\usepackage{multicol}
\usepackage{multirow}
\usepackage{makecell}
\usepackage{booktabs}

\newcolumntype{P}[1]{>{\centering\arraybackslash}p{#1}}
\newlength\savewidth

\title{Label Critic: Design Data Before Models}

\name{
\parbox{\textwidth}{
\centering
Pedro R. A. S. Bassi$^{1,2,3}$, Qilong Wu$^{1,4}$, Wenxuan Li$^1$,\\
Sergio Decherchi$^{2}$, Andrea Cavalli$^{2,3,5}$, Alan Yuille$^1$, Zongwei Zhou$^{1,*}\thanks{$^*$ Correspondence to: Zongwei Zhou (\href{mailto:zzhou82@jh.edu}{\texttt{\textbf{zzhou82@jh.edu}}})}$
}
}

\address{$^1$Johns Hopkins University \quad $^2$Italian Institute of Technology \quad $^3$University of Bologna \\ $^4$National University of Singapore \quad $^5$École Polytechnique Fédérale de Lausanne\\
{\small \texttt{\textbf{Code \& Data:}} \href{https://github.com/PedroRASB/Cerberus}{\texttt{\textbf{https://github.com/PedroRASB/LabelCritic}}}}
}

\begin{document}
%
\maketitle
\begin{abstract}


As medical datasets rapidly expand, creating detailed annotations of different body structures becomes increasingly expensive and time-consuming. We consider that requesting radiologists to create detailed annotations is unnecessarily burdensome and that pre-existing AI models can largely automate this process. Following the spirit \textit{don't use a sledgehammer on a nut}, we find that, rather than creating annotations from scratch, radiologists only have to review and edit errors if the Best-AI Labels have mistakes. To obtain the Best-AI Labels among multiple AI Labels, we developed an automatic tool, called \textbf{\pipeline}, that can assess label quality through tireless pairwise comparisons. Extensive experiments demonstrate that, when incorporated with our developed Image-Prompt pairs, pre-existing Large Vision-Language Models (LVLM), trained on natural images and texts, achieve 96.5\% accuracy when choosing the best label in a pair-wise comparison, without extra fine-tuning. By transforming the manual annotation task (30--60 min/scan) into an automatic comparison task (15 sec/scan), we effectively reduce the manual efforts required from radiologists by an order of magnitude. When the Best-AI Labels are sufficiently accurate (81\% depending on body structures), they will be directly adopted as the gold-standard annotations for the dataset, with lower-quality AI Labels automatically discarded. \pipeline\ can also check the label quality of a single AI Label with 71.8\% accuracy when no alternatives are available for comparison, prompting radiologists to review and edit if the estimated quality is low (19\% depending on body structures). 

\end{abstract}
%

\section{Introduction}\label{sec:introduction}

Publicly available abdominal CT datasets with per-voxel annotations have experienced rapid growth in recent years \cite{chou2024embracing,li2023medshapenet}. In 2020, datasets like KiTS \cite{heller2019kits19} and LiTS \cite{bilic2019liver} offered a few hundred annotated CT scans. By 2023, datasets such as AbdomenAtlas \cite{li2024well} and FLARE \cite{ma2022fast} expanded these scans significantly, now exceeding 10,000 annotated scans. This growth is enabled by AI-assisted annotation, where AI performs the initial segmentation and radiologists review and edit errors made by AI \cite{wasserthal2023totalsegmentator,qu2023annotating}. Despite AI assistance, the current scale---now with tens of thousands of annotations per dataset \cite{li2024abdomenatlas}---has made manual detection and editing of label errors increasingly impractical. This raises the question: \textit{Rather than having radiologists detect and edit AI errors, can we---again---use AI to automate these tasks and scale medical datasets?}

\begin{figure*}   
    \centering
    \includegraphics[width=0.96\textwidth]{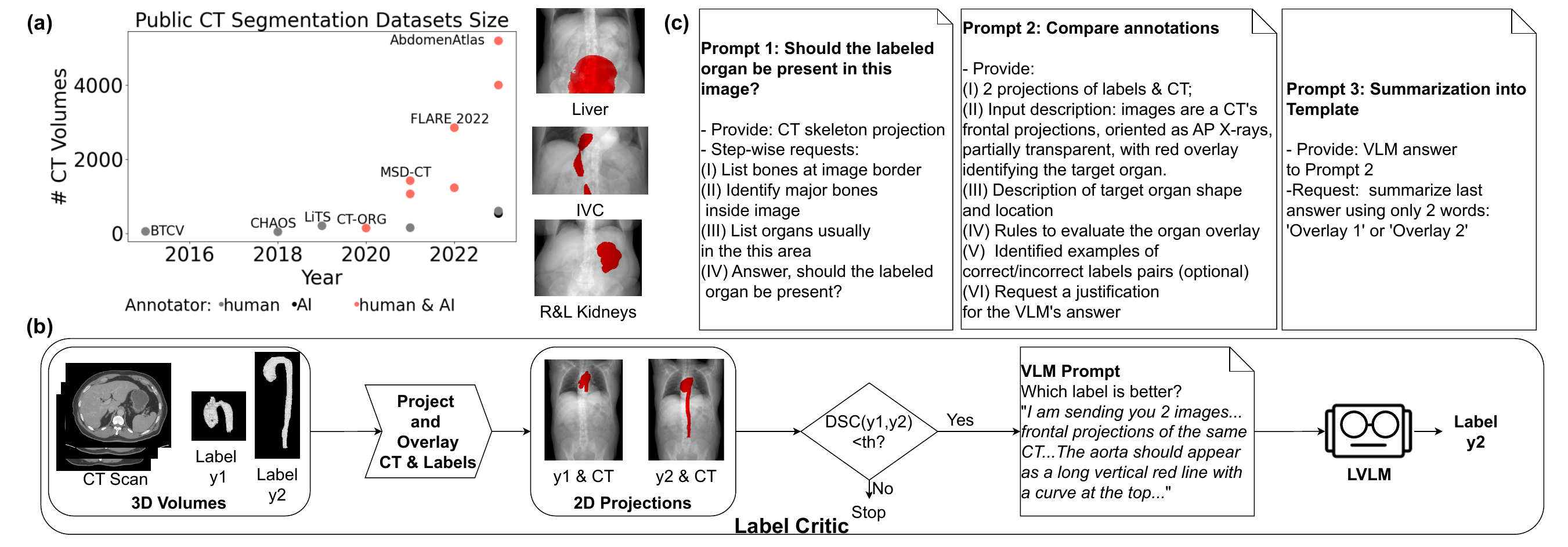}
    \label{fig:main_figure}
    \caption{\textbf{(a) Public CT datasets} with per-voxel labels are rapidly expanding, largely due to AI-assisted labeling. However, AI often makes obvious errors, exampled in the liver, IVC, and kidneys, highlighting the need for efficient, automated error detection. \textbf{(b) \pipeline\ pipeline for comparing labels.} (I) Frontally \textit{project} (\S\ref{sec:projections}) the CT scan and overlay it with the projections of two candidate labels (red), $y_{1}$ and $y_{2}$, creating two images; (II) \textit{verify the dice} score (DSC) between the 2 label projections, skip the comparison if DSC is above a class-specific threshold---avoiding comparing overly similar labels; (III) ask a \textit{LVLM} (\S\ref{sec:LVLMs}) to compare the labels and choose the most correct. If $y_{1}$ is a dataset label we are evaluating, we consider it wrong if the LVLM prefers $y_{2}$, the output of an alternative public segmentation model. \textbf{(c)}\textbf{ 3-Step Prompt Design.} \textit{Prompt 1} asks if the target organ should be in the CT, providing a skeleton projection as reference. If the LVLM says no, we select an empty label (if available) or flag the case for review. Otherwise, \textit{Prompt 2} asks the LVLM to compare two label overlays using class-aware prompts with anatomical guidance, optional in-context learning, and complexity based on the LVLM's background knowledge of each class (\S\ref{sec:prompt}). \textit{Prompt 3} asks the LVLM to summarizes its previous answer. Summarization provides an easily processable binary answer, but allows detailed justifications and step-by-step reasoning in earlier steps.}
    \label{fig:label-critic}
\end{figure*}

Automatic error detection is achievable most label errors in existing datasets because, simply put, \textit{critiquing is easier than creating}. This paper builds on two main insights. \textbf{First}, most errors made by AI are easy to detect\footnote{A common AI error in abdominal CT scans is mislabeling the aortic arch. This error is obvious, as the aorta should appear curved in its top, forming an arch (see Fig.~\ref{fig:label-critic}). Even non-experts can easily recognize such errors due to the aorta’s consistent size, position, and appearance across scans.} and do not require the time and expertise of busy, costly radiologists. \textbf{Second}, when multiple labels are available\footnote{The number of public AI models quickly raises \cite{antonelli2022medical}. Medical segmentation benchmarks provide diverse datasets, where participants train different architectures, providing a variety of labels for \pipeline\ to choose from. E.g., for abdominal organ segmentation in CT, we easily find solutions to the FLARE challenge  \cite{liu2023clip,huang2023revisiting,isensee2021nnu,liu2023flare}, 11 models trained on AbdomenAtlas are already public, and more will be released after Touchstone Benchmark \cite{li2024abdomenatlas}.}, comparing them to identify the highest-quality label is even simpler.

We discover that general-purpose Large Vision Language Models (LVLMs), like Llava and GPT-4V \cite{wang2024qwen2,liu2024visual}, trained on massive text-image datasets, can detect errors in medical datasets and compare the label quality among multiple label options \textit{without} additional fine-tuning. 
We present a new LVLM-based pipeline, \textbf{\pipeline}, which can effectively (1) detect a large portion (76.8\%) of the obvious label errors in existing medical datasets and (2) select the Best-AI Label by comparing multiple AI Labels.

We show that \pipeline\ can generalize to over 10,000 CT scans across 89 hospitals with minimal or no training data ($\leq$ 10). It detects 1,441 errors in the datasets, with overall accuracy of 96.5\% in detecting label errors and identify the Best-AI Labels in a pair-wise comparison. The success of \pipeline\ is attributed to our innovative \textbf{Input} and \textbf{Prompt} designs specialized for 3D CT scans and the integration of prior knowledge about body structures.

First, we design new inputs for LVLMs. Since most LVLMs are designed for 2D inputs, \pipeline\ uses 2D frontal projections of CT scans with transparent overlays of label projections, ensuring computational efficiency while preserving key volumetric information (\S\ref{sec:projections}). The projections resemble antero-posterior (AP) X-rays, making them familiar to general-purpose LVLMs.

Second, we design new prompts for LVLMs. They incorporate step-by-step guidance, anatomical descriptions, Dual Confirmation, and variable examples ranging from zero-shot to in-context learning with up to 10 label examples (\S\ref{sec:prompt}). This flexibility enables \pipeline\ to adapt quickly to new hospitals and segmentation classes, requiring few or no training samples, while avoiding overfitting to specific label error types (\S\ref{sec:prompt}, \S\ref{sec:LVLMs}). This is the first work to show LVLMs can compare semantic segmentations, using prior knowledge to choose the best AI model for each case and class.

\smallskip\noindent\textbf{\textit{Related Work.}}
Label quality control methods identify potential label errors by flagging uncertainty and inconsistency across AI models \cite{alba2018automatic,balcan2007margin,mahapatra2018efficient,scheffer2001active,nath2020diminishing}, but they do not specify which label is better, leaving radiologists to review each flagged case manually. In our dataset, this approach requires manual review of 4,348 labels across two AI models, a time-intensive task. Most existing QC methods are organ-specific (e.g., cardiac or muscle imaging \cite{tarroni2018learning,zaman2023segmentation}), limiting their scalability to other body structures. There is no prior methods leverage large vision-language models (LVLMs) for label quality control. Our LVLM-based method can significantly reduce manual workload for multi-organ segmentation\footnote{\label{organs}Spleen, gallbladder, pancreas, postcava, aorta, kidneys, spleen, and liver.} by comparing and selecting the best labels, discarding incorrect ones, and flagging only the most challenging cases for further manual review, streamlining the process efficiently.

\section{Methodology}

As shown in Fig.~\ref{fig:label-critic}, \pipeline\ includes projecting the CT scan and labels into 2D, calculating the Dice Similarity Coefficient (DSC) between labels, and prompting a LVLM to select the most accurate label. If the DSC is below a class-specific threshold—the dataset’s average class DSC minus one standard deviation—comparison is skipped, saving computational resources. The DSC check skips comparisons of labels with minor differences, focusing instead on substantial errors detectable through basic anatomical knowledge. When alternative labels or public segmentation models are unavailable, \pipeline\ assumes a non-comparative approach: it projects the CT with its single label and asks the LVLM to verify its anatomical accuracy, optionally using other CTs and labels as in-context examples.

\subsection{LVLM Architecture and (no) Training}
\label{sec:LVLMs}

Training AI for label error detection requires a dataset identifying both correct and incorrect labels, but assembling it is challenging: small medical datasets contain few errors, and finding errors in large AI-labeled datasets is labor-intensive--hence the need for automatic error detection. Although synthetic error generation is possible, training on either real or synthetic errors risks shortcut learning: models concentrating on the specific error types in the training dataset and failing to generalize well to unseen types \cite{Geirhos2020Shortcut}. To address these issues and enable broad adaptability across hospitals, we leverage zero-shot and few-shot learning. Given the limited per-voxel annotations in the training data of large vision-language models, robust out-of-distribution generalization is essential for our pipeline. We experimented with several LVLMs, selecting Qwen2-VL \cite{wang2024qwen2}---a large general-purpose model with 70 billion parameters and AWQ quantization for speed---because it can analyze multiple images per prompt, unlike alternatives such as LLaVA-7B \cite{liu2023improvedllava}, LLaVA-Med \cite{li2023llavamed}, and M3D-4B \cite{bai2024m3d}, which also yielded lower performance (see Tab.~\ref{tab:results}). Proprietary models like GPT-4V \cite{yang2023dawn} were not considered due to high API costs for processing large volumes of images.

\begin{table*}[t]
\caption{\textbf{\pipeline\ excels in two datasets.} We report Accuracy as the proportion of labels correctly evaluated out of the total evaluated. Each class contains an equal number of correct and incorrect labels. The LVLM used here is Qwen2-VL \cite{bai2023qwen}; we also tested Llava \cite{liu2024visual}, Llava-Med \cite{li2023llavamed}, and M3D \cite{bai2024m3d}, but these alternatives performed poorly, with average Accuracies of 54.1\%, 50.2\%, and 49.4\%, respectively, for error detection on AtlasBench.}
\centering
\scriptsize
\begin{tabular}{p{0.078\linewidth}p{0.058\linewidth}P{0.067\linewidth}P{0.067\linewidth}P{0.067\linewidth}P{0.067\linewidth}P{0.067\linewidth}P{0.067\linewidth}P{0.067\linewidth}P{0.067\linewidth}P{0.067\linewidth}}
    \toprule
     & & \multicolumn{9}{c}{AtlasBench (error detection)} \\
    \cmidrule{3-11}
    prompt & in-context & aorta & gallbladder & kidneys & liver & pancreas & postcava & spleen & stomach & average \\
    \midrule
    class-agnostic & 0-shot & 51.0 \scalebox{.55}{(530/1040)} & 50.0 \scalebox{.55}{(59/118)} & 84.9 \scalebox{.55}{(107/126)} & 55.6 \scalebox{.55}{(10/18)} & 63.2 \scalebox{.55}{(72/114)} & 0.0 \scalebox{.55}{(0/2)} & 40.0 \scalebox{.55}{(8/20)} & 66.7 \scalebox{.55}{(8/12)} & 54.8 \scalebox{.55}{(794/1450)} \\
    \midrule
    \multirow{3}{*}{class-aware} & 0-shot & 58.7 \scalebox{.55}{(610/1040)} & 50.8 \scalebox{.55}{(60/118)} & 89.7 \scalebox{.55}{(113/126)} & 83.3 \scalebox{.55}{(15/18)} & 85.1 \scalebox{.55}{(97/114)} & 50.0 \scalebox{.55}{(1/2)} & 80.0 \scalebox{.55}{(16/20)} & 50.0 \scalebox{.55}{(6/12)} & 63.3 \scalebox{.55}{(918/1450)} \\
     & 1-shot & 63.9 \scalebox{.55}{(665/1040)} & 50.8 \scalebox{.55}{(60/118)} & 83.3 \scalebox{.55}{(105/126)} & 83.3 \scalebox{.55}{(15/18)} & 76.3 \scalebox{.55}{(87/114)} & 100.0 \scalebox{.55}{(2/2)} & 70.0 \scalebox{.55}{(14/20)} & 50.0 \scalebox{.55}{(6/12)} & 65.8 \scalebox{.55}{(954/1450)} \\
     & 10-shot & 72.2 \scalebox{.55}{(751/1040)} & 50.8 \scalebox{.55}{(60/118)} & 77.0 \scalebox{.55}{(97/126)} & 83.3 \scalebox{.55}{(15/18)} & 80.7 \scalebox{.55}{(92/114)} & 100.0 \scalebox{.55}{(2/2)} & 75.0 \scalebox{.55}{(15/20)} & 75.0 \scalebox{.55}{(9/12)} & 71.8 \scalebox{.55}{(1041/1450)} \\
    \midrule
     & & \multicolumn{9}{c}{AtlasBench (label comparison)} \\
    \midrule
    class-agnostic & 0-shot & 78.7 \scalebox{.55}{(546/694)} & 68.0 \scalebox{.55}{(34/50)} & 95.7 \scalebox{.55}{(90/94)} & 100.0 \scalebox{.55}{(14/14)} & 97.1 \scalebox{.55}{(68/70)} & - \scalebox{.55}{(0/0)} & 100.0 \scalebox{.55}{(12/12)} & 100.0 \scalebox{.55}{(2/2)} & 81.8 \scalebox{.55}{(766/936)} \\
    \midrule
    class-aware & 0-shot & \textbf{96.5 \scalebox{.55}{(440/456)}}  & \textbf{74.4 \scalebox{.55}{(58/78)}}   & \textbf{96.4 \scalebox{.55}{(106/110)}} & \textbf{100.0 \scalebox{.55}{(12/12)}} & \textbf{92.2 \scalebox{.55}{(94/102)}} & \textbf{- \scalebox{.55}{(0/0)}}         & \textbf{100.0 \scalebox{.55}{(12/12)}} & \textbf{66.7 \scalebox{.55}{(4/6)}}     & \textbf{93.6 \scalebox{.55}{(726/776)}} \\
    \midrule
     & & \multicolumn{9}{c}{JHHBench (label comparison)} \\
    \midrule
    class-aware & 0-shot & \textbf{98.4 \scalebox{.55}{(1234/1254)}} & \textbf{92.9 \scalebox{.55}{(340/366)}} & \textbf{85.7 \scalebox{.55}{(12/14)}}   & \textbf{100.0 \scalebox{.55}{(62/62)}} & \textbf{100.0 \scalebox{.55}{(22/22)}} & \textbf{100.0 \scalebox{.55}{(346/346)}} & \textbf{100.0 \scalebox{.55}{(18/18)}} & \textbf{93.8 \scalebox{.55}{(122/130)}} & \textbf{97.5 \scalebox{.55}{(2156/2212)}} \\
    \bottomrule
\end{tabular}
\label{tab:results}
\end{table*}

\subsection{Projections and Overlays}
\label{sec:projections}

The usual 2D representation of CT volumes consists in their 2D slices. However, slices only show a small portion of the scan, and multiple slices would be needed to represent an annotation. Conversely, projections show through the entire body, conveying the entire CT and annotation in a single image. They cannot capture all possible label errors, such as holes inside annotations. However, they are a cost-effective solution: transformers' computational cost increases quadratically with input length, hindering the use of many CT slices as input. Antero-posterior projections of CT scans resemble AP X-rays, making them familiar and interpretable to pre-trained LVLMs. E.g., asked to describe the projections in Fig.~\ref{fig:label-critic}, GPT-4V says ``a frontal X-ray-like projection'' or ``a frontal projection of a CT scan''. Alg.~\ref{alg:projection} describes the projection procedure.

\begin{algorithm}
\caption{2D Projection of a 3D CT Scan}
\label{alg:projection}
\small
\begin{algorithmic}[1]
\STATE \textbf{Threshold:} Limit HU values to $[-500, 1500]$---a window that makes projections more X-ray-like.
\STATE \textbf{Project in 2D:} Sum over the antero-posterior axis.
\STATE \textbf{Normalize:} Apply $x_i = (x_i - x_{\min})/(x_{\max} - x_{\min})$.
\STATE \textbf{Resize \& RGB:} Resize to 512 p on the longest image side, keeping aspect ratio, and replicate in 3 channels (RGB).
\end{algorithmic}
\end{algorithm}

To project the label and overlay it over the CT projection, we first repeat Alg.~\ref{alg:projection} steps 1 and 2 for the label. Then, we zero the blue and green channels of the CT projection where the label projection is not 0, creating a semi-transparent red overlay that doesn’t obscure the CT. Also, we create skeleton projections to help the LVLM identifying missing or misplaced labels (see \S\ref{sec:prompt}). To create them, we use a window of [400, 2000] in Alg.~\ref{alg:projection} and enhance the projection's contrast with CLAHE \cite{pizer1987adaptive} (grid 8, clip 5) and gamma adjustment ($\gamma=0.6$). For M3D, we provide 3D CTs with labels overlaid in black (lowest CT HU value), as preliminary tests showed this color outperformed white or gray overlays, possibly due to its more natural look inside CTs.

\subsection{\textit{Prompt! Prompt! Prompt!}}\label{sec:prompt}

Prompt design impacts accuracy (\S\ref{sec:results}). In the large (N=5,195) AtlasBench CT dataset (\S\ref{sec:results}), we iteratively created a prompt, ran \pipeline, analyzed wrong LVLM answers, and improved the prompt accordingly. This process led to a standardized 3-step prompt, detailed in Fig.~\ref{fig:label-critic}.
Step 2, which requests the LVLM to compare two labels, is class dependent. Prompt complexity, strictness and number of in-context examples (from 0-10) depends on the LVLM background knowledge of the class: liver, spleen, kidneys and pancreas are classes the LVLM is more familiar with, allowing more complex and less strict prompts, with abstract shape references (e.g., “wedge-like”) and multiple anatomical landmarks (e.g.,“below the diaphragm”); VLMs are less familiar to aorta and postcava, and our prompts used simple anatomical descriptions and strict guidance, focusing on linear shape, extension, and continuity; stomach and gallbladder have less well-defined shape, and our prompt focus on label location and gross shape errors. For stomach we use in-context learning, providing one example of correct label.

We repeat Prompts 2 and 3 (Fig.~\ref{fig:main_figure}), inverting the image order in the LVLM input, and we check if its answers are consistent across the repetitions. This procedure, dubbed Dual Confirmation, reveals unreliable LVLM answers for minute label errors or cases where both labels are wrong. Also, we observed the LVLM itself can reject these comparisons, saying both labels are bad or similar. If Dual Confirmation finds inconsistent answers or the VLM rejects comparisons, we remove the case from the dataset, flagging for human review. To detect errors without label comparison, we skip the dice check and Dual Confirmation and modify prompt 2 to ask the LVLM to evaluate a single label, giving it examples of other CTs and correct/incorrect labels. We also created class-agnostic 3-step prompts, readily applicable to new classes, by removing class information in Prompt 2. Prompts were summarized for Llava, Llava-med and M3D, due to smaller context length. All prompts are available in our \href{https://github.com/PedroRASB/Cerberus}{code}.

\section{Results and Discussion}
\label{sec:results}

We created two datasets, \OurDatasetAbdomenAtlas\ and \OurDatasetJHH, to evaluate \pipeline. They contain errors from real public and private datasets, including mistakes in AI and human labels. As a ground truth, labels in \OurDatasetAbdomenAtlas\ and \OurDatasetJHH\ were manually deemed correct or incorrect. Both dataset have labels for eight abdominal organs\textsuperscript{\ref{organs}}.

\textbf{\OurDatasetAbdomenAtlas}: The public AbdomenAtlas dataset \cite{qu2023annotating,li2024abdomenatlas}, annotated by AI-assisted radiologists, includes 5,195 abdominal CT volumes from 88 hospitals worldwide. We used \pipeline\ to compare an intermediate development version of AbdomenAtlas (Beta) to the current release (1.0). Label Critic's DSC check (Fig.~\ref{fig:main_figure}) selected 1,450 labels with low DSC, finding labels that were updated from Beta to 1.0, potentially due to errors. We dubbed this subset \OurDatasetAbdomenAtlas. We have released it as the first public dataset specifically for benchmarking error detection and label comparison methods.

\textbf{\OurDatasetJHH}: JHH consists of 5,172 CT volumes from Johns Hopkins Hospital, annotated manually by radiologists. To construct \OurDatasetJHH, we compared these with pseudo-annotations from a public nnU-Net ResEncL trained on AbdomenAtlas during the Touchstone Benchmark \cite{isensee2021nnu}. Here, the Label Critic's DSC check selected 2,808 low-DSC labels. 

\textit{Label critic was accurate and generalized to diverse types of label errors.} For pair-wise comparison, it correctly chose the best label 97.5\% of the time in JHHBench, and 93.5\% in AtlasBench (Tab.~\ref{tab:results}). Most labels deemed wrong represent are AI errors, but \pipeline\ even uncovered 188 errors in human-made annotations--133 aorta errors due to the aortic arch falling out of the annotator's region of interest, and 55 label corruption cases like missing slices. We never trained the LVLMs in \pipeline, and its prompts were developed in AtlasBench considering AI errors only. Thus, finding errors in JHH represents strong OOD generalization.

\textit{AtlasBench results show easy adaptability to new classes, and demonstrate label comparisons are more accurate.}
While in-context learning (10-shot) improved non-comparative error detection, it still trailed behind the accuracy of the comparative Label Critic. Thus, even when only detecting errors in a dataset, one should prefer using Label Critic to compare the dataset's labels to outputs from public segmentation models. Label Critic’s class-tailored prompts outperformed class-agnostic ones, yet the latter still achieved 81.8\% accuracy, underscoring Label Critic’s efortless adaptability to new classes. 

The best-performing model in Label Critic was Qwen2-VL \cite{wang2024qwen2}, a large, general-purpose model with 70 billion parameters. It outperformed smaller models, including the Llava-7B \cite{liu2023improvedllava}, and medical fine-tuned models like Llava-Med \cite{li2023llavamed} and M3D \cite{bai2024m3d}, which both only reached chance-level accuracy (50\%). Qwen2-VL’s advantages come from its larger size and longer context length (32,768~vs.~8,000 tokens), which improve reasoning, handling instructions, and processing larger prompts. In contrast, Llava-Med and M3D, fine-tuned on smaller medical datasets (100K medical volumes for M3D~vs.~400M images for CLIP), struggled to generalize to \pipeline’s out-of-distribution (OOD) tasks \cite{uppaal2023finetuningneededpretrainedlanguage}, as label error detection and comparison are uncommon tasks in their training data.


\smallskip\noindent\textbf{\textit{Conclusion.}}
\pipeline\ proves to be highly effective for detecting and comparing label errors in organ segmentation, choosing the best label with an impressive accuracy of 97.5\% on JHHBench and 93.5\% on AtlasBench. This study is the first to show LVLMs can compare segmentation outputs and automatically select the Best-AI label for each sample, finding and discarding errors, and minimizing manual label revision. \pipeline\ generalized to multiple error types (both in AI and human labels) and it allows easy adaptation to new hospital and classes. Thus, it can help dataset creators enhance label quality in massive medical datasets, and model creators improve training data before training models. In future work, we plan to extend \pipeline\ to the tumor class.

\clearpage
\noindent\textbf{Acknowledgments.}
This work was supported by the Lustgarten Foundation for Pancreatic Cancer Research, the McGovern Foundation, and Istituto Italiano di Tecnologia (73010, Arnesano, LE, Italy; 16163, Genova, GE, Italy).

{\small
\bibliographystyle{IEEEbib}
\bibliography{refs,zzhou,wql}
}

\end{document}